\documentclass{esannV2}
\usepackage[dvips]{graphicx}
\usepackage[latin1]{inputenc}
\usepackage{amssymb,amsmath,array}
\usepackage{hyperref}
\usepackage{booktabs}
\usepackage{cleveref}
\usepackage{natbib}
\usepackage{xcolor}
\usepackage{floatrow} 
\usepackage{microtype}
\hypersetup{
            pdftitle={Four Quadrants of Difficulty:\\ A Simple Categorisation and its Limits},
            pdfauthor={Vanessa Toborek; Sebastian M\"uller; Christian Bauckhage},
            pdfkeywords={curriculum learning, nlp, sample difficulty},
            bookmarksnumbered=true, 
            pdfcreator={Vanessa Toborek},
            pdfproducer={LaTeX (customized)},
            }

\begin{document}
\title{Four Quadrants of Difficulty:\\ A Simple Categorisation and its Limits}

\author{Vanessa Toborek$^{1,2}$, Sebastian M\"uller$^{1,2}$ and Christian Bauckhage$^{1,2,3}$
%
\vspace{.3cm}\\
%
$^{1}$University of Bonn, Germany $^{2}$Lamarr Institute, Germany \\
$^{3}$Fraunhofer Institute IAIS, Germany
}

\maketitle

\begin{abstract}
Curriculum Learning (CL) aims to improve the outcome of model training by estimating the difficulty of samples and scheduling them accordingly.
In NLP, difficulty is commonly approximated using task-agnostic linguistic heuristics or human intuition, implicitly assuming that these signals correlate with what neural models find difficult to learn. We propose a four-quadrant categorisation of difficulty signals -- human vs.\ model and task-agnostic vs.\ task-dependent -- and systematically analyse their interactions on a natural language understanding dataset. We find that task-agnostic features behave largely independently and that only task-dependent features align. These findings challenge common CL intuitions and highlight the need for lightweight, task-dependent difficulty estimators that better reflect model learning behaviour. 
\end{abstract}

\section{Introduction}
\label{sec:intro}
Curriculum Learning (CL) is based on the intuition that neural network training should be structured in ways that mimic human learning: starting from easier concepts before progressing to harder ones \citep{Elman1993,Bengio2009}. 
Any CL strategy therefore consists of two components: a function that assigns a difficulty score to each training instance and a scheduler that determines when each instance becomes available during training. 
In NLP, estimating difficulty is particularly challenging because linguistic difficulty is multi-dimensional and hard to capture with a single measure \citep{Battisti2020}. As a result, CL research relies on a wide range of task-agnostic measures -- such as sentence length, syntactic complexity, or readability scores -- as well as approaches that draw directly on human intuition \citep{Elgaar2023, Toborek2025a}. 

Implicitly, these strategies assume that linguistic difficulty as perceived by humans aligns with what is actually difficult for neural networks. 
Yet, several domains such as psycholinguistics, annotation disagreement \citep{Plank2022}, and training-dynamics analysis \citep{Swayamdipta2020} reflect distinct notions of difficulty, each grounded in different assumptions about human processing, task ambiguity, or model learning behaviour.
Despite the accumulating evidence, the field lacks a systematic analysis of how linguistic difficulty, human disagreement, and model learning difficulty relate. While existing work sometimes distinguishes task-specific from task-agnostic difficulty \citep{Christopoulou2022,Toborek2025b}, this distinction remains purely operational and does not aim to capture the broader conceptual space.

To address this gap, we introduce a principled, four-quadrant classification of difficulty that distinguishes (i) human vs.\ model sources and (ii) task-agnostic vs.\ task-dependent information. This classification synthesises insights from readability research, human label variation, and training dynamics, and enables us to formulate and empirically test expectations how different difficulty signals interact.
We perform a systematic, cross-quadrant analysis across (1) task-agnostic human linguistic features, (2) task-dependent human difficulty, (3) task-agnostic model signals, and (4) task-dependent model difficulty. 
Our results show that task-agnostic difficulties behave orthogonally to 
task-dependent difficulties. 
Linguistic complexity fails to predict annotation disagreement or model learning difficulty; only task-dependent human and task-dependent model signals show meaningful alignment. These findings challenge a central assumption behind many heuristic CL strategies: that task-agnostic linguistic difficulty unilaterally captures the difficulty a model experiences during training.

The implications are twofold. First, the success of task-agnostic CL heuristics must stem from mechanisms other than accurate difficulty estimation, such as distributional reshaping during training by the CL scheduler. Second, annotation entropy, while highly informative, remains expensive. Our results therefore underscore the need for developing new, inexpensive ways to approximate task-dependent difficulty at pre-processing time, before model training begins.

\section{A Categorisation of Difficulty Signals} 
\label{sec:taxonomy}
\begin{table}[]
    \centering
    \resizebox{1.008\textwidth}{!}{%
    \begin{tabular}{p{2.5cm}p{1cm}p{8cm}}
    \toprule
         Task-agnostic & Human & Length, word rarity, SLE, diversity, complexity, FRE, age-of-acquisition, concreteness, prevalence \\
         Task-agnostic & Model & Perplexity \\
         Task-dependent & Human & Inter-annotator disagreement \\
         Task-dependent & Model & Confidence, variability, correctness, loss \\
    \bottomrule
    \end{tabular}
    }
    \caption{Overview of all sample difficulty proxies. We propose a categorisation in four distinct groups.}
    \label{tab:overview-proxies}
\end{table}
\textbf{The Four Quadrants of Difficulty}
We categorise difficulty signals into four quadrants defined by two dimensions: their source (human vs.\ model) and their scope (task-agnostic vs.\ task-dependent). Table~\ref{tab:overview-proxies} provides an overview.
\newline\textit{Task-agnostic Human Difficulty (TA-H)} This group comprises measures intended to capture the linguistic difficulty of an input independently of any downstream task. 
Some of these proxies are simple surface-level heuristics, such as average sentence length or word rarity, computed with respect to frequency distributions in the training corpus. The Flesch-Reading-Ease (FRE) score, originally designed for longer texts, is frequently applied to sentence-level readability assessment \citep{Battisti2020}. We include the psycholinguistic measures age-of-acquisition (AOA), concreteness, and prevalence, which have been shown to be informative predictors of lexical complexity \citep{Desai2021}. Further, we incorporate two syntactic measures: diversity, defined as the set size of part-of-speech tags in the input, and complexity, defined as the average depth of a sentence dependency parse tree. Finally, we also test the learned, reference-less metric SLE that has been shown to correlate well with human perception of difficulty \citep{Cripwell2023}.
\newline\textit{Task-agnostic Model Difficulty (TA-M)} In most machine learning settings, model-based difficulty signals cannot be obtained without task-specific training. In NLP, however, large pre-trained language models offer a way to approximate task-agnostic model difficulty by examining the behaviour of the model prior to finetuning. Perplexity reflects how well a pre-trained model predicts the input under its learned language distribution, and therefore captures aspects of fluency and lexical expectation derived from pre-training. We use the average perplexity over masked tokens in the input as a task-agnostic model signal.
\newline\textit{Task-dependent Human Difficulty (TD-H)} 
For this category, we consider inter-annotator disagreement. Given multiple annotations per instance, disagreement reflects human uncertainty about the correct label. Such uncertainty may arise from heterogeneous sources, including lexical or syntactic complexity, inherent semantic ambiguity, underspecification in the input, or subjective annotator variation \citep{Plank2022}. We quantify disagreement using annotation entropy, a label-distribution-based measure of uncertainty.
\newline\textit{Task-dependent Model Difficulty (TD-M)} When training a model on a specific task, one can derive task-dependent difficulty proxies from its training dynamics \citep{Swayamdipta2020}. We track several statistics: the model's average confidence in the correct label, its correctness across epochs, and its variability, defined as the standard deviation of confidence across training. We additionally monitor the mean and standard deviation of the loss. These metrics provide a post hoc view of how difficult each instance is for the model to learn.
\medskip
\newline
\textbf{Expected Interactions Between Quadrants}
The four-quadrant classification allows us to 
motivate expectations about how different difficulty signals should relate to each other:
\newline
\textit{H1: Internal coherence of TA-H.} Linguistic difficulty is known to be multi-dimensional, with lexical, syntactic, and conceptual complexity capturing distinct aspects. Hence, we expect low internal correlation among TA-H features.
\newline
\textit{H2: TA-H $\leftrightarrow$ TA-M.} Perplexity reflects a pre-trained language model's surprisal over an input sequence, driven by its learned lexical and syntactic expectations. Because several TA-H features partially relate to lexical predictability, we expect moderate correlations between perplexity and human linguistic features.
\newline
\textit{H3: TA-H $\leftrightarrow$ TD-H.} Building on observations that linguistic complexity may contribute to human disagreement \citep{Plank2022}, we expect moderate correlations between linguistic difficulty and annotation entropy.
\newline
\textit{H4: TA-H $\leftrightarrow$ TD-M.} Given that CL strategies using surface linguistic features have proven effective in some settings, we expect interactions between linguistic difficulty and model learning difficulty.
\newline
\textit{H5: TD-H $\leftrightarrow$ TD-M.} Prior work on ``dataset cartography'' shows that instances with high label ambiguity tend to be harder for models to learn \citep{Swayamdipta2020}. We expect a positive relationship between human disagreement and model training difficulty.

Although these hypotheses are grounded in prior work, the field lacks a systematic evaluation of whether these theoretically motivated interactions hold in practice. This classification serves as the conceptual foundation for our empirical study.

\section{Empirical Study: Testing the Quadrant Interactions}
\label{sec:empirical-study}
\begin{figure}
    \centering
    \includegraphics[width=\textwidth]{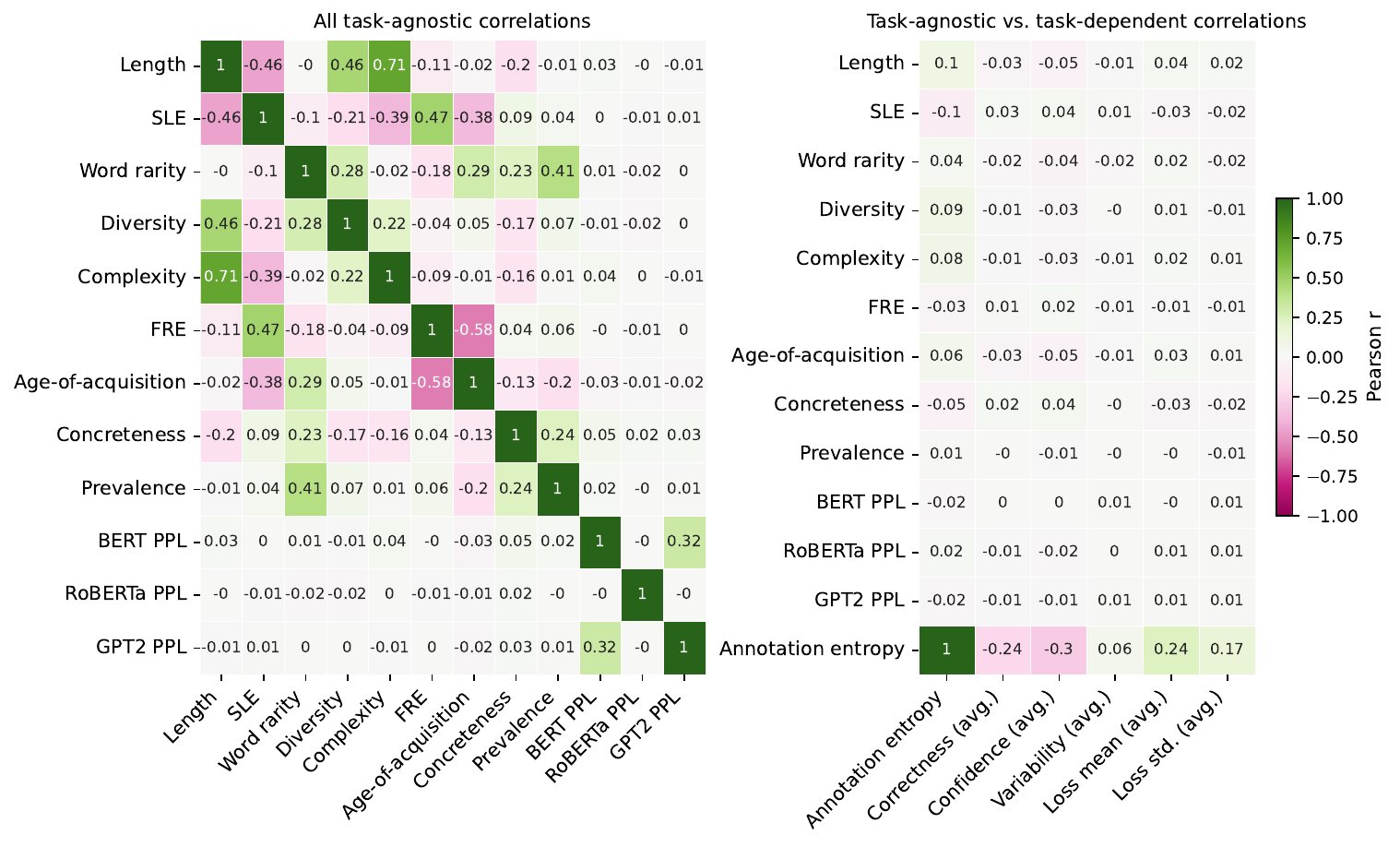}
    \caption{Pearson correlation for the four difficulty quadrants. (Left) Correlation among task-agnostic human and task-agnostic model difficulty signals. (Right) Correlation for all task-agnostic difficulty signals with all task-dependent ones. All TD-M are averaged over all three models and ten random seeds each.}
    \label{fig:TA-corr}
\end{figure}

\begin{figure}
    \centering
    \includegraphics[width=\linewidth]{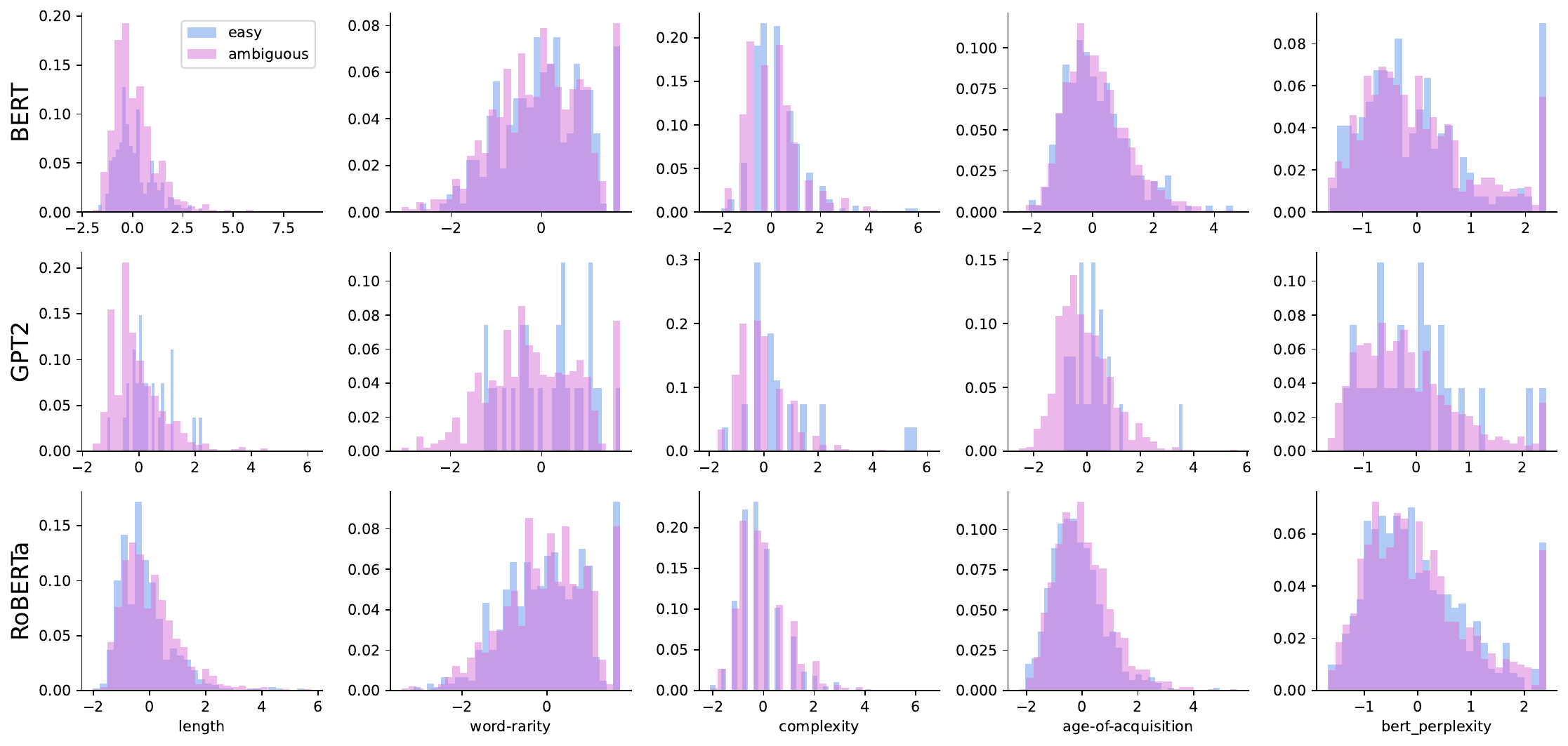}
    \caption{Histograms of easy vs. ambiguous samples (based on the respective model confidence and variability) for selected task-agnostic proxies: length, word rarity, complexity, age of acquisition, and BERT perplexity.}
    \label{fig:histograms}
\end{figure}

We empirically evaluate our hypotheses on the interactions between difficulty quadrants using the SNLI dataset \citep{Bowman2015}. It provides four independent annotator labels for each instance in the training split. This allows for computing annotation entropy as the TD-H difficulty proxy. 
TA-H features are calculated in a preprocessing step. 
For TA-M features we compute perplexities using pre-trained BERT-base \citep{Devlin2019}, RoBERTa-base \citep{Liu2019}, and GPT-2-base \citep{Radford2019} models. 
TD-M features are collected for 3\,647 data points at twelve evenly spaced checkpoints during finetuning. BERT/RoBERTa are trained with batch size $b$=$64$ and learning rate $lr$=3$\times$$10^{-5}$, GPT-2 with $b$=$16$ and $lr$=$10^{-5}$.
All models use $5$ epochs, AdamW (weight decay $0.01$), and a linear $lr$ schedule with $6\%$ warm-up.

We evaluate the expected relationship between quadrants in four ways: (1) we compute the Pearson correlation (a) within the TA-H quadrant, (b) between TA-H and TD-H, and (c) between TA-H and TD-M. (2) To assess whether all TA features jointly predict either TD difficulty, we perform multivariate regression using either TD-H or TD-M features as targets. (3) Following the diagnosing approach of dataset cartography \citep{Swayamdipta2020}, we identify the top and bottom 25\% of easy-to-learn and ambiguous data points. We then compare the distribution on task-agnostic features to test whether they meaningfully separate them. (4) As a sanity check, we replicate the established relationship between annotation entropy and model ambiguity \citep{Swayamdipta2020} for the different model architectures.

Together, these analyses provide a comprehensive evaluation of whether difficulty signals across the four quadrants show the hypothesised interaction.
\medskip\newline\textbf{Results} 
\Cref{fig:TA-corr} (left) shows the Pearson correlations among task-agnostic difficulty signals. Supporting \textit{H1}, TA-H correlations are mostly moderate ($r<0.5$) to low ($r<0.2$), with only length-complexity ($r=0.71$) and AOA-FRE ($r=-0.58$) standing out. This indicates that TA-H signals capture distinct facets of linguistic difficulty rather than a single underlying factor. 
Turning to \textit{H2}, the bottom of \Cref{fig:TA-corr} (left) shows that perplexity does not correlate with TA-H difficulty signals, providing no evidence for the hypothesised weak alignment.
\Cref{fig:TA-corr} (right) extends this picture: all TA features show virtually no correlation with TD-H (\textit{H3}) or any TD-M signal (\textit{H4}). This suggests that (a) SNLI label disagreement is likely driven by inherent ambiguity and less by linguistic complexity, and (b) surface-level linguistic properties are poor predictors of what models find difficult to learn.
To test these relationships multivariately, we regress each task-dependent signal on all TA features using both linear and tree-based models. 
Predictive power remains negligible ($R^2<0.05$ for annotation entropy and $R^2<0.1$ for all TD-M metrics), reinforcing the weak link between TA-H and TD difficulty. 
Distributional analysis of ``easy'' vs.\ ``ambiguous'' instances, defined via dataset cartography, provides further evidence. As shown in \Cref{fig:histograms}, the distributions of task-agnostic features overlap almost entirely for both groups across all models, indicating that linguistic difficulty does not separate easy-to-learn from ambiguous samples. 
Finally, the bottom row of \Cref{fig:TA-corr} (right) confirms the expected correlation structure among task-dependent signals (\textit{H5}): annotation entropy aligns with correctness, confidence, and average loss, reflecting their shared dependence on label uncertainty.

\section{Lessons for Curriculum Learning}
\label{sec:lessons-for-cl}
Our analysis for the SNLI dataset shows that task-agnostic and task-dependent difficulty signals behave largely independently, challenging the common, implicit assumption that linguistic difficulty, as captured by shallow proxies, directly approximates model learning difficulty.
Yet, both types of difficulty measures have been reported to yield successful curricula in practice, indicating that CL effectiveness cannot be attributed to difficulty estimation alone. Instead, it emerges from interactions between difficulty measures, scheduling decisions, and task characteristics, with scheduler design, particularly how difficulty is enforced and exposure evolves over time, playing a central role.
We further confirm the task-dependent model difficulty aligns with annotator entropy, supporting human disagreement as a meaningful difficulty signal. However, its high acquisition cost limits practical use. A key direction for future work is to develop lightweight, pre-computed approximations of task-dependent difficulty that better reflect the model's learning behaviour and enable more principled curriculum design.

\begin{footnotesize}
\bibliographystyle{unsrt}
\bibliography{references}

@inproceedings{Plank2022,
  author       = {B. Plank},
  title        = {{The ``Problem'' of Human Label Variation: On Ground Truth in Data, Modeling
                  and Evaluation}},
  booktitle    = {{EMNLP}},
  year         = {2022},
  url          = {https://doi.org/10.18653/v1/2022.emnlp-main.731},
  doi          = {10.18653/V1/2022.EMNLP-MAIN.731},
}

@inproceedings{Swayamdipta2020,
  author       = {S. Swayamdipta and
                  R. Schwartz and
                  N. Lourie and
                  Y. Wang and
                  H. Hajishirzi and
                  N. A. Smith and
                  Y. Choi},
  title        = {{Dataset Cartography: Mapping and Diagnosing Datasets with Training
                  Dynamics}},
  booktitle    = {{EMNLP}},
  year         = {2020},
  url          = {https://doi.org/10.18653/v1/2020.emnlp-main.746},
  doi          = {10.18653/V1/2020.EMNLP-MAIN.746},
}

@inproceedings{Bengio2009,
  title     = {{Curriculum Learning}},
  author    = {Y. Bengio and J. Louradour and R. Collobert and J. Weston},
  booktitle = {ICML},
  year      = 2009,
  url = {https://dl.acm.org/doi/10.1145/1553374.1553380}
}

@article{Elman1993,
  title        = {{Learning and Development in Neural Networks: The Importance of Starting Small}},
  author       = {{J. L. Elman}},
  volume       = 48,
  number       = 1,
  url          = {https://doi.org/10.1016/0010-0277(93)90058-4},
  year         = 1993,
  journal = {Cognition}
}

@inproceedings{Elgaar2023,
    title = {{HuCurl: Human-induced Curriculum Discovery}},
    author = "M. Elgaar and H. Amiri",
    booktitle = "ACL",
    year = "2023",
    url = "https://aclanthology.org/2023.acl-long.104/",
    doi = "10.18653/v1/2023.acl-long.104",
}

@inproceedings{Toborek2025a,
    title = {{Beyond Shallow Heuristics: Leveraging Human Intuition for Curriculum Learning}},
    author = {V. Toborek  and
      S. M{\"u}ller  and
      T. Selbach  and
      T. Horv{\'a}th and
      C. Bauckhage},
    booktitle = "ICNLSP",
    year = "2025",
    url = "https://aclanthology.org/2025.icnlsp-1.10/",
}

@inproceedings{Christopoulou2022,
    title = {{Training Dynamics for Curriculum Learning: A Study on Monolingual and Cross-lingual NLU}},
    author = "F. Christopoulou and G. Lampouras  and I. Iacobacci",
    booktitle = "EMNLP",
    year = "2022",
    url = "https://aclanthology.org/2022.emnlp-main.167/",
}

@inproceedings{Desai2021,
    title = {{LCP-RIT at {S}em{E}val-2021 Task 1: Exploring Linguistic Features for Lexical Complexity Prediction}},
    author = "A. T. Desai  and
      K. North  and
      M. Zampieri  and
      C. Homan",
    booktitle = "SemEval-2021",
    year = "2021",
    url = "https://aclanthology.org/2021.semeval-1.67/",
    doi = "10.18653/v1/2021.semeval-1.67",
}

@inproceedings{Toborek2025b,
  author       = {V. Toborek and
                  F. Seiffarth and
                  S. M{\"{u}}ller and
                  T. Horv{\'{a}}th and
                  C. Bauckhage},
  title        = {{Exploring Curriculum Learning for Languages: Lessons from Regular
                  Language Tasks}},
  booktitle    = {Discovery Science},
  year         = {2025},
  url          = {https://doi.org/10.1007/978-3-032-05461-6\_37},
  doi          = {10.1007/978-3-032-05461-6\_37},
}

@inproceedings{Battisti2020,
	title        = {{A Corpus for Automatic Readability Assessment and Text Simplification of German}},
	author       = {A. Battisti and D. Pf\"utze and A. S\"auberli and M. Kostrzewa and S. Ebling},
	year         = 2020,
	booktitle    = {{LREC}},
	url          = {https://aclanthology.org/2020.lrec-1.404/},
}

@inproceedings{Cripwell2023,
    title = {{Simplicity Level Estimate (SLE): A Learned Reference-Less Metric for Sentence Simplification}},
    author = {L. Cripwell  and J. Legrand and C. Gardent},
    booktitle = "EMNLP",
    year = "2023",
    url = "https://aclanthology.org/2023.emnlp-main.739/",
    doi = "10.18653/v1/2023.emnlp-main.739",
}

@inproceedings{Bowman2015,
    title = "A large annotated corpus for learning natural language inference",
    author = "S. R. Bowman and G. Angeli  and C. Potts  and C. D. Manning",
    booktitle = "EMNLP",
    year = "2015",
    url = "https://aclanthology.org/D15-1075/",
    doi = "10.18653/v1/D15-1075",
}

@inproceedings{Devlin2019,
    title = {{BERT: Pre-training of Deep Bidirectional Transformers for Language Understanding}},
    author = "J. Devlin  and
      M.-W. Chang  and
      K. Lee  and
      K. Toutanova",
    booktitle = "ACL",
    year = "2019",
    url = "https://aclanthology.org/N19-1423/",
    doi = "10.18653/v1/N19-1423",
}

@article{Radford2019,
  title={{Language models are unsupervised multitask learners}},
  author={A. Radford and J. Wu and R. Child and D. Luan and D. Amodei and I. Sutskever},
  journal={OpenAI blog},
  year={2019}
}

@article{Liu2019,
  author       = {Y. Liu and
                  M. Ott and
                  N. Goyal and
                  J. Du and
                  M. Joshi and
                  D. Chen and
                  O. Levy and
                  M. Lewis and
                  L. Zettlemoyer and
                  V. Stoyanov},
  title        = {{RoBERTa: A Robustly Optimized BERT Pretraining Approach}},
  journal      = {CoRR},
  year         = {2019},
}
\end{footnotesize}

\end{document}